\documentclass[10pt,twocolumn,letterpaper]{article}

\usepackage[pagenumbers]{cvpr}   

\definecolor{cvprblue}{rgb}{0.21,0.49,0.74}
\usepackage[pagebackref,breaklinks,colorlinks,allcolors=cvprblue]{hyperref}
\usepackage{siunitx}
\usepackage{cuted}
\usepackage{tabularx}

\usepackage[most]{tcolorbox}

\newtcolorbox{alqaPromptBox}[1]{%
    enhanced,
    breakable,
    colback=black!1.5,
    colframe=black!35,
    colbacktitle=black!8,
    coltitle=black,
    title={#1},
    fonttitle=\bfseries\footnotesize,
    boxrule=0.45pt,
    arc=0.7mm,
    left=1.7mm,
    right=1.7mm,
    top=1.1mm,
    bottom=1.1mm,
    before skip=5pt,
    after skip=6pt
}

\def\confName{CVPR}
\def\confYear{2027}

\title{Automating Multi-View Quality Assessment of 3D Annotation Layouts}

\author{
Hongyi Yang$^{1}$ \quad
Jinao Wu$^{1}$ \quad
Yanwei Zhang$^{1}$ \quad
Shengwen Fu$^{1}$ \\
Haibin Ling$^{2}$ \quad
Bingyao Huang$^{1}$ \quad
Jingwei Qu$^{1}$
\thanks{\raggedright
Corresponding author: Jingwei Qu
(\texttt{qujingwei@swu.edu.cn}).}
 \\[4pt]
$^{1}$College of Computer and Information Science,
Southwest University, Chongqing, China \\
$^{2}$Intelligent Computing and Application Lab,
Westlake University, Zhejiang, China
}

\begin{document}
\maketitle

\begin{strip}
\centering
\includegraphics[width=\linewidth]{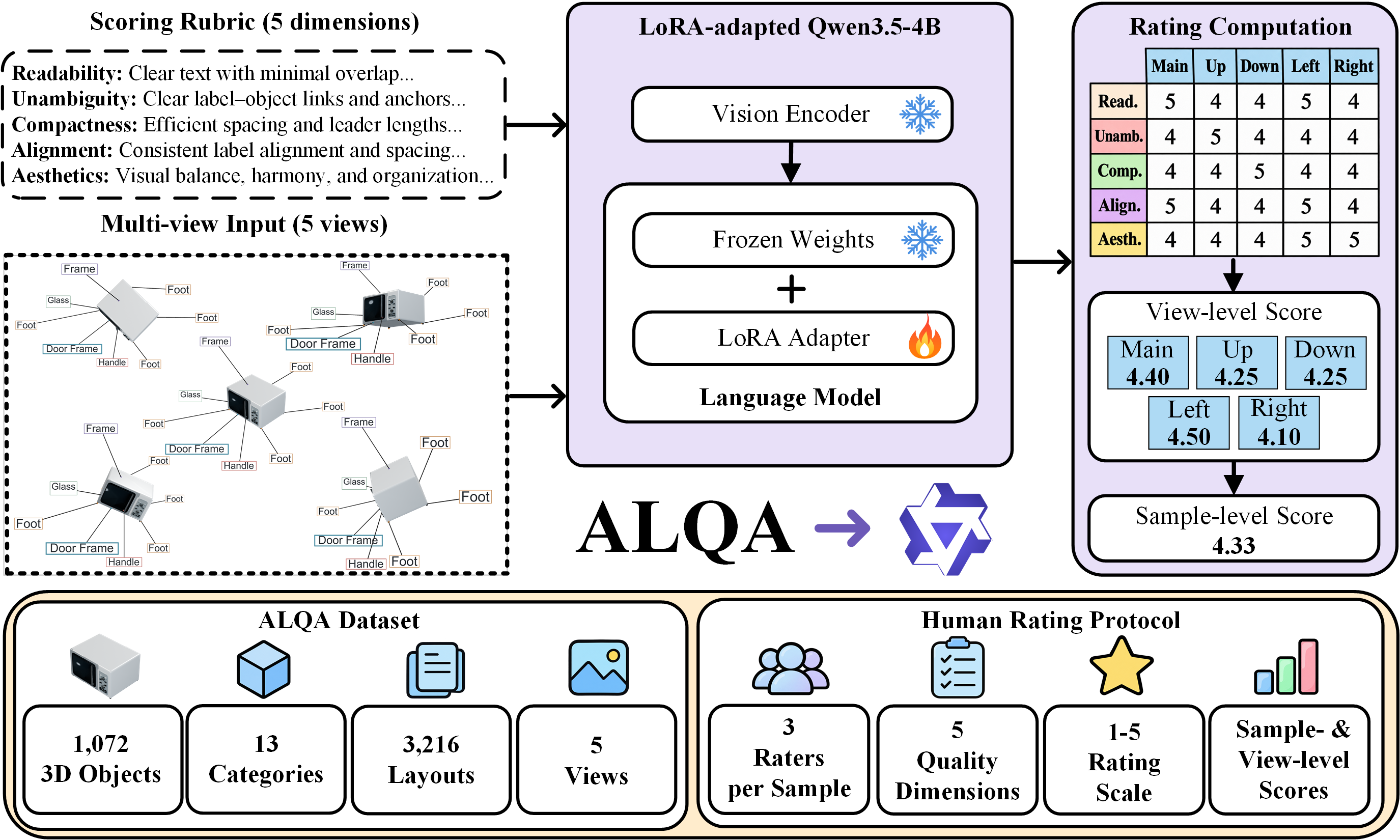}
\captionsetup{hypcap=false}
\captionof{figure}{Overview of our MLLM-based method and the ALQA (Annotation Layouts with Quality Assessment) dataset for multi-view quality assessment of 3D annotation layouts.}
\captionsetup{hypcap=true}
\label{fig:teaser}
\end{strip}

\begin{abstract} 
Text labels provide descriptive information about elements in visual scenes. In 3D visualization, the arrangement of these labels and their connections to target locations forms an annotation layout. Assessing layout quality supports comparisons between alternative layouts. However, existing quality measures often focus on specific layout properties and may not fully reflect human judgments of overall quality. Moreover, a layout may appear clear from one viewpoint but suffer from occlusion or clutter from another. To address these issues, we propose a multi-view quality assessment method based on a multimodal large language model (MLLM). We fine-tune a compact MLLM using structured supervision from human ratings to predict a rating for each quality dimension in each view. These ratings are aggregated into view-level and sample-level quality scores. To support training and evaluation, we construct the Annotation Layouts with Quality Assessment (ALQA) dataset, containing 3,216 layout samples from 1,072 3D objects across 13 categories. Each layout is rendered from 5 predefined viewpoints, and each view is independently rated by 3 human raters across 5 quality dimensions. On ALQA, our method agrees more closely with human reference scores than the evaluated MLLM baselines, and a user study demonstrates strong agreement with human preferences in layout ranking.
\end{abstract}


\section{Introduction}
\label{sec:intro}
In the field of visualization, text annotations, also called labels, provide descriptive information about elements in visual scenes. In 3D visualization and extended reality applications, they generally help users identify objects and their parts~\cite{bekos2019external,Borhani2024TVCG}. An anchor marks a target location on the object, and a leader-line connects it to the corresponding label. The arrangement of labels, anchors, and leader-lines forms an annotation layout. Existing studies mainly focus on layout generation and optimization~\cite{tatzgern2014hedgehog,vcmolik2018real,vcmolik2020mixed,chen2024rl,bobak2024reinforced,qu2025graph}. By contrast, automatic quality assessment of 3D annotation layouts remains under-investigated. 

Existing 3D annotation layout methods often use predefined quality measures as optimization objectives or evaluation metrics, such as label occlusion~\cite{chen2024rl} and leader-line length~\cite{vcmolik2018real}. These measures capture specific layout properties and help improve usability. However, layouts that satisfy basic label readability and label--target association requirements may still differ in visual organization and aesthetics. Predefined measures may not fully reflect human judgments of these differences, limiting their ability to guide layout comparisons according to human preferences. Moreover, assessing 3D annotation layouts requires accounting for viewpoint-dependent differences. A layout may appear clear from one viewpoint but suffer from occlusion or clutter from another. Multi-view representations provide complementary visual information about 3D objects~\cite{Hamdi2021ICCV}, motivating their use in assessing layout quality across viewpoints.

Visual quality assessment spans photographic aesthetics, graphic design, and video quality. Earlier work introduced human-rated datasets and models for photographic aesthetics and image quality~\cite{murray2012ava,ma2017alamp,talebi2018nima}. Research on graphic design also includes preference-based evaluation of generated posters~\cite{lai2026posterreward}. More recently, multimodal large language models (MLLMs) have been used for quality scoring~\cite{wu2024qalign,you2025deqa} and to provide textual analyses alongside scores~\cite{cao2026artimuse,jia2026vital}. However, comprehensive quality assessment of 3D annotation layouts remains underexplored.

Motivated by these observations, we propose an MLLM-based method for quality assessment of 3D annotation layouts (\cref{fig:teaser}). We fine-tune a compact MLLM using structured supervision from human ratings to assess layouts across multiple viewpoints and quality dimensions. The MLLM predicts ratings for each view and dimension, which are aggregated into view-level and sample-level quality scores. Public datasets with substantial numbers of 3D annotation layouts remain scarce. To support training and evaluation, we construct the Annotation Layouts with Quality Assessment (ALQA) dataset. Inspired by recent datasets that adopt 5-view settings to improve object coverage~\cite{Wang2024CVPR,Fan2025CVPR}, we render each layout from 5 predefined viewpoints. Each view is rated by 3 human raters across 5 quality dimensions.


In summary, our main contributions include:
\begin{itemize}
    \item We propose the first MLLM-based method for multi-view quality assessment of 3D annotation layouts. It predicts ratings for each view and dimension, then aggregates them into view-level and sample-level quality scores.

    \item We construct ALQA, the first human-rated dataset containing 3,216 layout samples from 1,072 3D objects. Each sample includes 5 rendered views and human ratings across 5 quality dimensions, supporting the training and evaluation of automatic quality assessment methods.

    \item Experiments on ALQA show that our method outperforms the evaluated MLLM baselines, and a user study demonstrates strong agreement with human preferences.
\end{itemize}


\section{Related Work}
\label{sec:related}
\subsection{Annotation Layout Design}
Annotation layout design has been widely studied in 3D visualization and extended reality applications~\cite{bekos2019external,Borhani2024TVCG}. Traditional methods mainly rely on geometric constraints and rule-based optimization to organize labels and reduce visual conflicts~\cite{tatzgern2014hedgehog,vcmolik2018real,vcmolik2020mixed}. They maintain temporal coherence under viewpoint changes~\cite{tatzgern2014hedgehog}, optimize label placement in ghosted 3D views~\cite{vcmolik2018real}, or combine internal and external labels to reduce overlap and ambiguous associations~\cite{vcmolik2020mixed}. More recent methods explore reinforcement learning (RL) for dynamic AR label placement and point-feature labeling~\cite{chen2024rl,bobak2024reinforced}, while graph-based methods model interactions among labels and incorporate visual features~\cite{qu2025graph,zhang2026mixture}. Beyond optimization, human-centered studies examine how annotation design affects visual search and noticeability~\cite{kruijff2018influence} and identify broader design considerations from annotation practice~\cite{rahman2026designing}.

Existing studies evaluate generated layouts using metrics for label occlusion, leader-line crossings, and label movement~\cite{chen2024rl}, comparisons with reference layouts~\cite{qu2025graph,zhang2026mixture}, and expert evaluations or user studies~\cite{vcmolik2020mixed,bobak2024reinforced}. These evaluations focus on specific layout properties or require human judgments. However, automatic assessment that reflects human judgments of the overall quality of 3D annotation layouts remains underexplored.

\subsection{Visual Quality Assessment}
Visual quality assessment has been extensively explored for photographs and graphic designs, with different methods focusing on perceptual and structural quality. For photographs, AVA provides a benchmark of human aesthetic ratings~\cite{murray2012ava}. Subsequent methods incorporate image composition and holistic layout~\cite{mai2016composition,ma2017alamp}, model rating distributions~\cite{talebi2018nima}, or generate textual critiques~\cite{chang2017aesthetic}. Graphic design assessment considers both spatial organization and visual aesthetics. Design-o-Meter learns to evaluate graphic design quality and suggests modifications to improve visual appeal~\cite{goyal2025designometer}, while Agentic-DRS assesses alignment, composition, aesthetics, and color~\cite{nag2026agentic}. For posters, quality assessment places greater emphasis on typography and readability. PosterReward learns a reward model from preference data, focusing on typography and layout~\cite{lai2026posterreward}, whereas PosterIQ benchmarks layout, typography, readability, and overall quality~\cite{feng2026posteriq}. Advertising image assessment addresses additional perceptual aspects, with A3 evaluating perceptual attention, formal interest, and desire impact~\cite{ji2026a3}. These studies inform the selection of quality dimensions for annotation layout assessment.

MLLMs have been applied to image quality, image aesthetics, graphic design, and video quality assessment, supporting both numerical scoring and textual interpretation~\cite{wu2024qinstruct,wu2024qalign,huang2024aesexpert,you2025deqa,wang2025aigv,cao2026artimuse,jia2026vital,patnaik2025aesthetiq}. For image quality, methods leverage human feedback or score distributions to improve score prediction. Q-Instruct enhances low-level visual understanding through instruction tuning with human natural language feedback~\cite{wu2024qinstruct}, while DeQA-Score models score distributions to improve quality score regression~\cite{you2025deqa}. Image aesthetic assessment incorporates human judgments at different levels of detail and can provide textual analyses alongside scores. AesExpert uses feedback ranging from coarse aesthetic grades to fine-grained descriptions~\cite{huang2024aesexpert}, whereas ArtiMuse jointly predicts aesthetic scores and generates textual analyses~\cite{cao2026artimuse}. For graphic design, AesthetiQ combines overlap-based metrics with MLLM-based aesthetic comparisons~\cite{patnaik2025aesthetiq}. AIGV-Assessor extends quality assessment to video by predicting quality scores and pairwise preferences across multiple perceptual dimensions~\cite{wang2025aigv}. Other methods develop unified frameworks for visual assessment. Q-Align uses discrete textual quality levels to predict human-aligned scores across image quality, image aesthetics, and video quality~\cite{wu2024qalign}, while VITAL unifies quality scoring and interpretation across images and videos~\cite{jia2026vital}.

Recent studies explore MLLMs for spatial understanding across views and frames~\cite{yang2025thinking,jeong2026multiview,xu2026multispatial,li2026starr1}. Building on MLLM-based visual assessment, our work focuses on the viewpoint-dependent quality of 3D annotation layouts. We use structured supervision from human ratings to predict a rating for each quality dimension in each view.

\section{Methodology}
\label{sec:method}
\begin{figure*}[t]
    \centering
    \includegraphics[width=\textwidth]{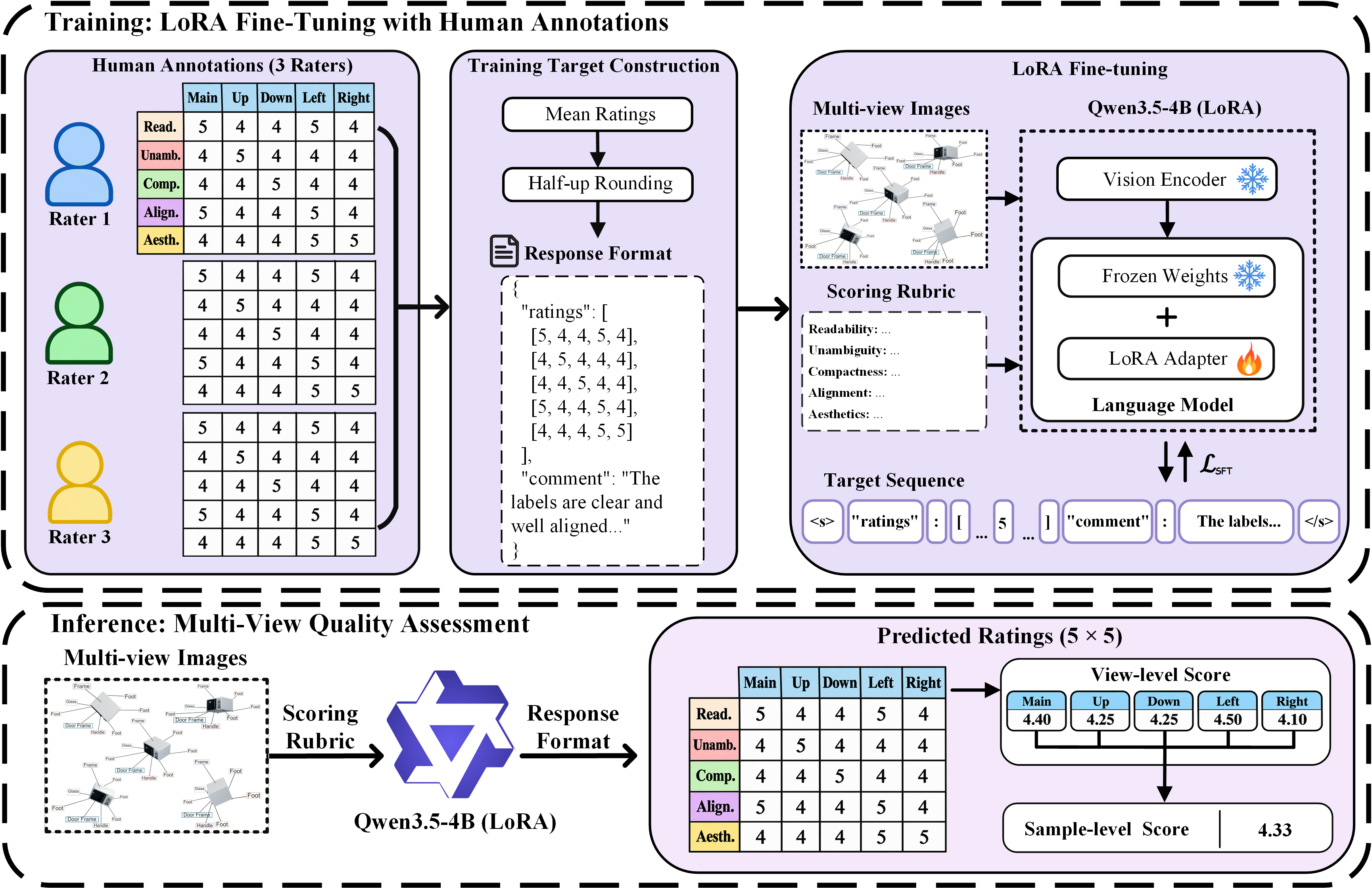}
    \caption{Overview of the proposed multi-view quality assessment pipeline. During training, we construct structured targets from human ratings and comments to fine-tune an MLLM with LoRA. During inference, predicted ratings for each view and quality dimension are aggregated into view-level and sample-level quality scores.}
    \label{fig:method_overview}
    \vspace{-3mm}
\end{figure*}

Our method formulates 3D annotation layout quality assessment as a structured multi-view rating task. Given 5 rendered views of a layout, an MLLM fine-tuned with low-rank adaptation (LoRA)~\cite{hu2022ICLR} predicts an integer rating for each quality dimension in each view. These ratings are aggregated into view-level and sample-level quality scores. An overview of the pipeline is shown in~\cref{fig:method_overview}.

\subsection{Problem Formulation}
\label{sec:problem}
In this formulation, each sample corresponds to an annotation layout that specifies the arrangement of labels, anchors, and leader-lines for a 3D object. To account for differences in layout quality across viewpoints, each layout is rendered from 5 predefined viewpoints, $\mathbb{V}=\{\texttt{main},\allowbreak \texttt{up},\allowbreak \texttt{down},\allowbreak \texttt{left},\allowbreak \texttt{right}\}$. The rendered images are provided to the model in the order listed. Layout quality is assessed along 5 dimensions, $\mathbb{D}=\{\texttt{Readability},\allowbreak \texttt{Unambiguity},\allowbreak \texttt{Compactness},\allowbreak \texttt{Alignment},\allowbreak \texttt{Aesthetics}\}$.

\noindent\textbf{Input:} For a sample $i$, the 5 rendered images are represented by $\mathbb{I}_i=\{\mathbf{I}_{i,v}\}_{v\in\mathbb{V}}$, where $\mathbf{I}_{i,v}$ is the image rendered from viewpoint $v$. The 5 images are provided together with a textual scoring rubric.

\noindent\textbf{Output:} For each viewpoint $v\in\mathbb{V}$ and quality dimension $d\in\mathbb{D}$, the model predicts an integer rating
$\hat{s}_{i,v,d}\in\{1,\ldots,5\}$. These ratings form a $5\times5$ matrix $\hat{\mathbf{S}}_i=[\hat{s}_{i,v,d}]_{d\in\mathbb{D},\,v\in\mathbb{V}}$. The model generates a textual response containing these ratings and a short sample-level comment, which is not used for quantitative scoring. The view-level quality score is computed as
\begin{equation}
\label{eq:view_score}   
\hat{q}_{i,v}=\sum_{d\in\mathbb{D}} w_d \hat{s}_{i,v,d},
\end{equation}
where the dimension weights $w_d$ are $0.25$, $0.25$, $0.25$, $0.15$, and $0.10$, following the order in which the dimensions are listed in $\mathbb{D}$. We assign greater weights to Readability, Unambiguity, and Compactness to emphasize the functional aspects of layout quality. Aesthetics receives the lowest weight, reflecting its more subjective nature. The sample-level quality score is
\begin{equation}
\label{eq:sample_score}
\hat{Q}_i=\sum_{v\in\mathbb{V}} \alpha_v \hat{q}_{i,v},
\end{equation}
where the viewpoint weight $\alpha_v$ is $0.40$ for $v=\texttt{main}$ and $0.15$ otherwise. The main view receives greater weight as the primary viewing direction, while the other 4 views are weighted equally to capture quality differences across viewpoints. These aggregation weights are predefined and remain fixed across all samples in our experiments. They can be adjusted to reflect the relative importance of quality dimensions and viewpoints in different tasks.

\subsection{Multi-view Quality Assessment}
To capture differences in layout quality across viewpoints and dimensions, we construct structured training targets from human ratings and comments. These targets supervise ratings for each quality dimension in each view, preserving detailed quality information that cannot be represented by a single aggregated score.

\noindent\textbf{Supervised Fine-tuning.} 
We perform supervised fine-tuning (SFT) of a pretrained MLLM using LoRA, with adapters applied to the attention and feed-forward projections of the language model. All pretrained parameters remain frozen, including those of the vision encoder. Each image is preceded by a marker identifying its viewpoint, and a textual scoring rubric specifies the 5 quality dimensions and their rating criteria.

\noindent\textbf{Structured Supervision.} 
For sample $i$, the human reference rating $\bar{y}_{i,v,d}$ is the mean rating across raters for viewpoint $v$ and quality dimension $d$. To match the model's integer rating scale of 1--5, we round the human reference ratings using the half-up rounding function
\begin{equation}
\mathcal{H}(s)=\min\left(5,\max\left(1,\left\lfloor s+0.5\right\rfloor\right)\right).
\label{eq:half_up}
\end{equation}
The resulting integer rating is $\tilde{y}_{i,v,d}=\mathcal{H}(\bar{y}_{i,v,d})$. The target response contains these 25 ratings and a short sample-level comment derived from the raters' comments.

\noindent\textbf{Training Objective.} 
We optimize the LoRA parameters by minimizing the causal language modeling loss
\begin{equation}
\mathcal{L}_{\mathrm{SFT}}=-\sum_i\sum_{t=1}^{T_i}\log p_{\theta,\phi}\left(z_{i,t}\mid\mathbb{I}_i,R,z_{i,<t}\right),
\end{equation}
where $p_{\theta,\phi}$ denotes the model's conditional token distribution, $\theta$ and $\phi$ denote the frozen pretrained parameters and trainable LoRA parameters, respectively, and $R$ denotes the textual scoring rubric. For sample $i$, the target response is tokenized as $\mathbf{z}_i=(z_{i,1},\ldots,z_{i,T_i})$, where $z_{i,<t}$ denotes the preceding tokens and $T_i$ is the number of tokens in sample $i$'s target response. During inference, we extract the predicted ratings from the generated response and aggregate them using the predefined dimension and viewpoint weights into view-level and sample-level quality scores, as specified in~\cref{eq:view_score,eq:sample_score}.

\section{ALQA Dataset}
\label{sec:dataset}
We construct ALQA, a dataset of 3D annotation layouts with human quality ratings. It contains 3,216 samples constructed from 1,072 3D objects across 13 categories, with 3 layouts for each object. Each sample is rendered from 5 predefined viewpoints and independently rated by 3 raters with backgrounds in visualization or design across 5 quality dimensions. The dataset is constructed through layout generation, multi-view rendering, and human annotation. The supplemental material provides the dataset construction pipeline, category statistics, and an example sample.

\subsection{Dataset Construction}
We select the 3D objects from PartNet~\cite{Mo2019CVPR} and use their part-level semantic annotations to define the labels in the layouts. Each label displays the name of an object part on a thin rectangular panel in 3D space. An anchor is a 3D point on the corresponding part mesh, and a leader-line connects the anchor to the label panel. Across the selected objects, we identify 9,872 labels from 130 semantic classes. Each object has 1 to 24 labels, averaging 9.21. 

For each object, we use distinct generation strategies to construct 3 layouts of varying quality. Layout 1 is manually constructed. Layout 2 is generated using a rule-based method. Layout 3 is derived from Layout 2 through random position swaps and perturbations of selected labels.

We render each layout from the 5 predefined viewpoints in $\mathbb{V}$ using virtual cameras with the same perspective projection settings at a resolution of $750\times500$. The main camera is oriented along the normalized $[1,1,1]$ direction, while the 4 auxiliary cameras are positioned $45^\circ$ away from the main viewing direction, toward the \texttt{up}, \texttt{down}, \texttt{left}, and \texttt{right} directions of the main image plane, respectively. All cameras remain at the same distance from the object center and point toward it. Borhani~\etal note that larger viewpoint changes may introduce occlusion and perspective distortion in 3D annotation displays~\cite{Borhani2024TVCG}. Accordingly, we use a moderate offset to provide viewpoint variation while keeping the views within a reasonable observation range. The 5 views provide complementary visual information for assessing layout issues across viewpoints, including occlusion, leader-line crossings, spatial dispersion, and local misalignment.

\subsection{Human Annotation}
We develop a multidimensional rating scheme for 3D annotation layout quality. Prior studies have examined a range of factors in annotation layout design, including text legibility, label--object correspondence, spatial arrangement, leader geometry, and visual organization~\cite{Hartmann2005SG,Gotzelmann2005CompAesth,Vollick2007UIST,Cmolik2010CAG}. Building on these considerations, we select the 5 complementary quality dimensions in $\mathbb{D}$. Their definitions are described in~\cref{tab:quality_dimensions}.

\begin{table}[t]
    \caption{Definitions of the 5 layout quality dimensions.}
    \label{tab:quality_dimensions}
    \centering
    \small
    \begin{tabularx}{\linewidth}{l>{\raggedright\arraybackslash}X}
        \toprule
        {Dimension} & {Definition} \\
        \midrule
        Readability & Visibility and legibility of label text, considering overlap and occlusion. \\

        Unambiguity & Clarity of label–object correspondence through clear leaders and anchors. \\

        Compactness & Spatial efficiency in terms of proximity, leader length, dispersion, and crowding. \\

        Alignment & Regularity of local label groups, including rows, columns, edge alignment, and spacing. \\

        Aesthetics & Global visual quality, considering balance, harmony, clutter, and overall organization. \\
        \bottomrule
    \end{tabularx}
    \vspace{-3mm}
\end{table}

We recruit 3 researchers with visualization or design backgrounds to independently assess all samples. Each rater evaluates every sample across all 5 viewpoints. For each view, the 5 quality dimensions are rated on a 1--5 scale. For each sample, the interface displays all 5 views simultaneously, with 1 view active for rating and the others available as visual context. Raters can switch between views and zoom in for closer inspection. Raters are not informed of the layout generation strategies or other raters' ratings. We obtain 9,648 complete sample--rater records.

We apply the same weighted aggregation as in~\cref{eq:view_score,eq:sample_score} to the rounded human reference ratings $\tilde{y}_{i,v,d}$ to obtain the human reference scores $q_{i,v}$ and $Q_i$ at the view and sample levels, respectively. Individual ratings are retained for further analysis. Each rater also provides a short comment. We deduplicate and merge the 3 comments into a sample-level comment that is excluded from quantitative scoring. We assess inter-rater reliability using the intraclass correlation coefficient (ICC) under a two-way random-effects model with absolute agreement. The ICC is 0.599 for a single rater and 0.818 for the mean of 3 raters, indicating greater reliability when scores are averaged across raters.

\begin{figure}[!t]
    \centering
    \includegraphics[width=\columnwidth]{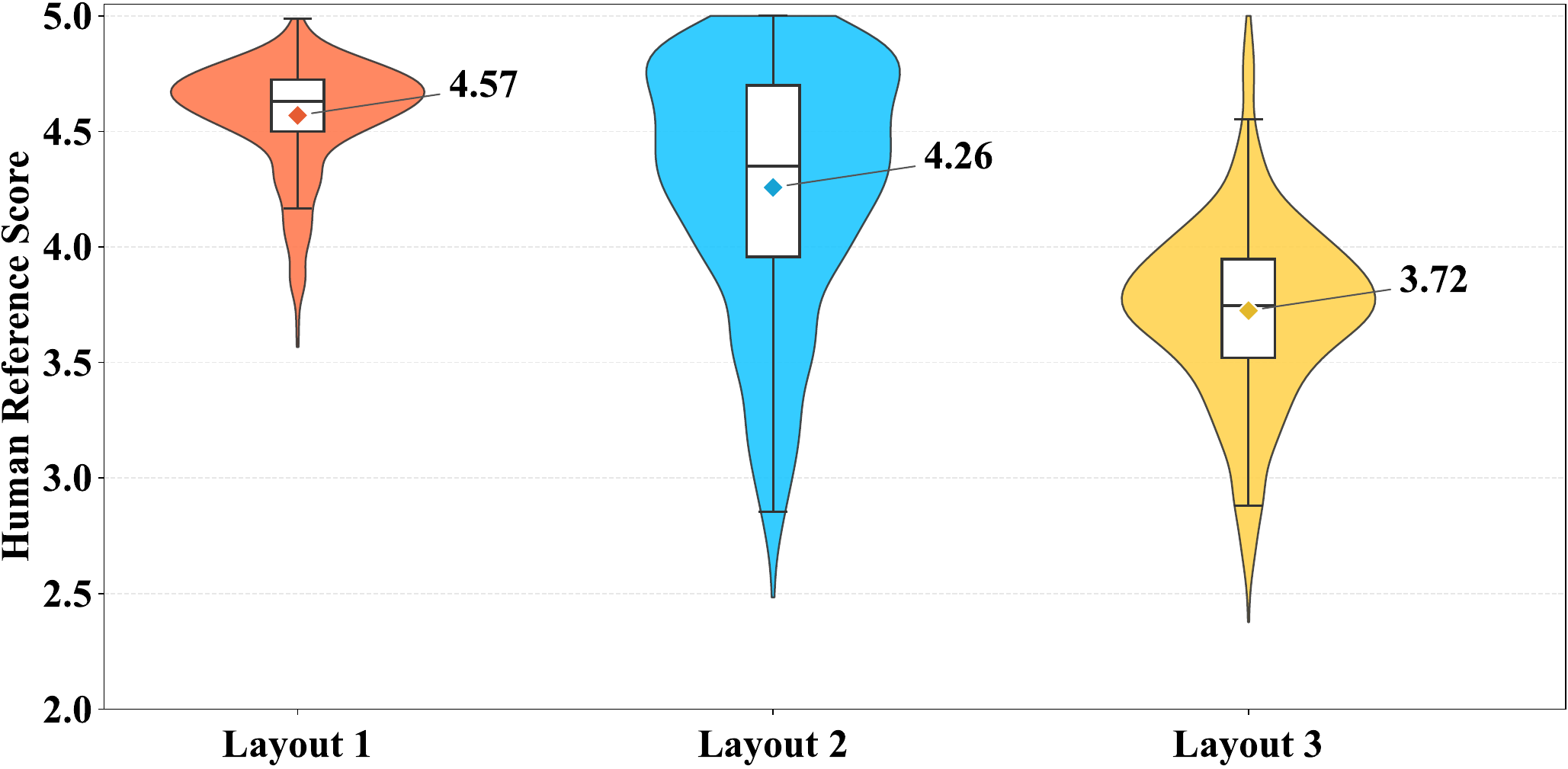}
    \caption{Distribution of sample-level human reference scores for Layouts 1, 2, and 3. Diamonds indicate mean scores.}
    \label{fig:quality_distribution}
    \vspace{-3mm}
\end{figure}

\subsection{Quality Distribution and Splits}
The raw human ratings span the full 1--5 scale, and 78.74\% are 4 or 5. This concentration reflects in part the limited presence of severely degraded layouts, which offer little practical value. Objects with excessive numbers of labels are also uncommon, as dense labeling can reduce visual clarity and make the information harder to interpret. At the dimension level, using the same view weights as in the sample-level aggregation, the mean raw ratings for Readability, Unambiguity, Compactness, Alignment, and Aesthetics are 4.26, 4.04, 4.10, 4.04, and 4.10, respectively. Across all samples, the mean sample-level human reference score is 4.18. As shown in~\cref{fig:quality_distribution}, Layouts 1, 2, and 3 exhibit distinct score distributions, with means of 4.57, 4.26, and 3.72, respectively. To examine the relationship between label count and layout quality, we average the reference scores of the 3 layouts for each object. Objects with 1--5 labels have a mean score of 4.52, compared with 3.35 for those with more than 20 labels. Using Spearman's rank correlation~\cite{spearman1904proof}, we find a strong negative association between label count and layout quality ($\rho=-0.872$, $p<0.001$).

We split ALQA into training, validation, and test sets at an approximate ratio of 7:1:2, containing 750, 107, and 215 objects, respectively. The split is performed at the object level, ensuring that all 3 layouts of the same object remain in the same subset. During splitting, we jointly consider object category, label count, and quality distribution to keep the 3 subsets well matched. We will release the dataset and trained model weights upon acceptance.

\section{Experiments}
\label{sec:experiments}
We evaluate our method on ALQA through comparisons with prompt-only and fine-tuned MLLM baselines, ablation studies, and a user study.

\subsection{Experimental Setup}
\noindent\textbf{Implementation Details.}
We use Qwen3.5-4B~\cite{QwenTeam2026Qwen35} as the default backbone and fine-tune it with LoRA using rank 16, $\alpha=32$, and dropout 0.05. We use AdamW with an initial learning rate of $2\times10^{-4}$ and a linear learning rate schedule. Fine-tuned models are trained for up to 3 epochs, and the checkpoint with the lowest validation loss is selected. The computing hardware includes 4 NVIDIA GeForce RTX 5090 GPUs with 32~GB memory each and 2 Intel Xeon Gold 6530 CPUs. Fine-tuning of the default Qwen3.5-4B model is performed on a single GPU and takes 6.25 hours. Multiple GPUs are used only when required for fine-tuning larger models such as Qwen3.5-27B.

\noindent\textbf{Evaluation Metrics.}
We evaluate our method using mean absolute error (MAE)~\cite{willmott2005advantages} and quadratic weighted Cohen's kappa (QWK)~\cite{cohen1968weighted,fleiss1973equivalence} at the sample and view levels. The 4 metrics are Sample MAE (S-MAE), View MAE (V-MAE), Sample QWK (S-QWK), and View QWK (V-QWK). Lower MAE and higher QWK indicate better performance. For $N$ test samples with predicted sample-level scores $\hat{Q}_i$ and human reference scores $Q_i$, S-MAE is
\vspace{-2mm}
\begin{equation}
\vspace{-2mm}
\mathrm{S\text{-}MAE}=\frac{1}{N}\sum_{i=1}^{N}\left|\hat{Q}_i-Q_i\right|.
\end{equation}
V-MAE is computed analogously using the predicted and human reference view-level scores, $\hat{q}_{i,v}$ and $q_{i,v}$. For S-QWK, we apply the half-up rounding function in~\cref{eq:half_up} to both $\hat{Q}_i$ and $Q_i$, converting them to ordinal levels from 1 to 5. For reference level $a$ and predicted level $b$, $o_{ab}$ denotes the observed frequency of their paired occurrences, and $e_{ab}$ denotes the expected frequency under independence. The quadratic disagreement weight is $\omega_{ab}=(a-b)^2/(K-1)^2$, where $K=5$. QWK is computed as
\vspace{-2mm}
\begin{equation}
\vspace{-2mm}
\mathrm{QWK}=1-\frac{\sum_{a=1}^{K}\sum_{b=1}^{K}\omega_{ab}o_{ab}}{\sum_{a=1}^{K}\sum_{b=1}^{K}\omega_{ab}e_{ab}}.
\end{equation}
V-QWK is computed in the same way using the corresponding view-level scores.

\subsection{Evaluation Results}
\noindent\textbf{Prompt-Only MLLMs.}
We compare our method with 4 prompt-only MLLM baselines, namely Qwen3.8-Max~\cite{QwenTeam2026Qwen38Max}, GPT-6 Astra~\cite{OpenAI2026GPT6Astra}, Claude Opus 5.5~\cite{Anthropic2026ClaudeOpus55}, and Gemini 4 Argon~\cite{GoogleDeepMind2026Gemini4Argon}. As shown in~\cref{tab:general_mllm_comparison}, our method outperforms all baselines across the 4 metrics. Compared with Qwen3.8-Max, the strongest baseline, S-MAE decreases from 0.2614 to 0.1388, a reduction of 46.9\%, while S-QWK increases from 0.7601 to 0.8645, a gain of 0.1044. Although Qwen3.8-Max achieves the best performance among the prompt-only baselines, it remains behind our method, suggesting that prompt-only MLLMs may not fully capture human judgments of annotation layout quality. Compared with GPT-6 Astra, Claude Opus 5.5, and Gemini 4 Argon, our method reduces S-MAE by 74.3\%, 82.7\%, and 83.2\%, respectively, with corresponding S-QWK gains of 0.4180, 0.4573, and 0.4898. The improvement also extends to view-level assessment, with the lowest V-MAE of 0.1770 and the highest V-QWK of 0.8246. These results highlight the effectiveness of fine-tuning on human ratings for assessment at both the sample and view levels.

\begin{table}[t]
    \centering
    \caption{Comparison with prompt-only MLLMs on ALQA.}
    \label{tab:general_mllm_comparison}
    \small
    \begin{tabular}{lcccc}
        \toprule
        Model & S-MAE & S-QWK & V-MAE & V-QWK \\
        \midrule
        Qwen   & 0.2614 & 0.7601 & 0.3717 & 0.6918 \\
        GPT    & 0.5403 & 0.4465 & 0.5798 & 0.4967 \\
        Claude & 0.8014 & 0.4072 & 0.8141 & 0.4294 \\
        Gemini & 0.8282 & 0.3747 & 0.8450 & 0.3999 \\
        \midrule
        Ours   & \textbf{0.1388} & \textbf{0.8645} & \textbf{0.1770} & \textbf{0.8246} \\
        \bottomrule
    \end{tabular}
    \vspace{-3mm}
\end{table}

\begin{table}[t]
    \centering
    \caption{Comparison with fine-tuned MLLMs on ALQA.}
    \label{tab:open_weight_comparison}
    \setlength{\tabcolsep}{3pt}
    \resizebox{\linewidth}{!}{
    \begin{tabular}{lcccccS[table-format=2.2,mode=text,reset-text-family=false,reset-text-series=false]}
        \toprule
        Model & Params. & S-MAE & S-QWK & V-MAE & V-QWK & {Time(s)} \\
        \midrule
        MiniCPM-V & 1.30B & 0.1486 & 0.8592 & 0.1859 & 0.8118 & \bfseries 1.18 \\
        Gemma     & 4.50B & 0.1542 & 0.8631 & 0.1950 & 0.8211 & 3.41 \\
        Ministral & 3.85B & 0.1584 & 0.8422 & 0.1956 & 0.8226 & 11.21 \\
        Granite   & 4.00B & 0.1867 & 0.8329 & 0.2392 & 0.7834 & 3.12 \\
        InternVL & 4.73B & 0.1978 & 0.8040 & 0.2559 & 0.7397 & 5.93 \\
        LLaVA     & 4.74B & 0.1562 & 0.8456 & 0.1931 & 0.8096 & 5.58 \\
        Phi       & 5.57B & 0.1731 & 0.8193 & 0.2109 & 0.7997 & 15.59 \\
        \midrule
        Ours      & 4.66B & \textbf{0.1388} & \textbf{0.8645} & \textbf{0.1770} & \textbf{0.8246} & 2.80 \\
        \bottomrule
    \end{tabular}
    }
    \vspace{-3mm}
\end{table}

\noindent\textbf{Fine-Tuned MLLMs.}
We further compare our method with 7 open-weight MLLMs: MiniCPM-V 4.6~\cite{OpenBMB2026MiniCPMV46}, Gemma 4 E4B IT~\cite{GemmaTeam2026Gemma4}, Ministral 3 3B~\cite{Liu2026Ministral3}, Granite Vision 4.1 4B~\cite{IBM2026GraniteVision41}, InternVL3.5-4B~\cite{Wang2025InternVL35}, LLaVA-OneVision-1.5-4B-Instruct~\cite{An2025LLaVAOneVision15}, and Phi-4-multimodal-instruct~\cite{Abouelenin2025Phi4}. All models are fine-tuned and evaluated using the same protocol as our method. We report the average inference time for each test sample. LLaVA-OneVision-1.5-4B-Instruct and Phi-4-multimodal-instruct use native Transformers inference due to limited support in our vLLM environment. As shown in~\cref{tab:open_weight_comparison}, our method achieves the best results across all 4 metrics, with an S-MAE of 0.1388 and an S-QWK of 0.8645. Compared with MiniCPM-V 4.6 and Gemma 4 E4B IT, our method reduces S-MAE by 6.6\% and 10.0\%, while improving S-QWK by 0.0053 and 0.0014, respectively. The smaller differences in S-QWK, particularly for Gemma 4 E4B IT, indicate that similar ordinal agreement can coexist with differences in continuous score accuracy. Compared with LLaVA-OneVision-1.5-4B-Instruct and Ministral 3 3B, our method reduces S-MAE by 11.1\% and 12.4\%, respectively. Larger reductions of 19.8\%, 25.7\%, and 29.8\% are observed against Phi-4-multimodal-instruct, Granite Vision 4.1 4B, and InternVL3.5-4B. These differences among models of comparable parameter scales show that backbone size alone does not determine assessment accuracy. MiniCPM-V 4.6 achieves competitive performance with only 1.30B parameters and the fastest inference time of 1.18~s, compared with 2.80~s for our method. At the view level, our method also achieves the lowest V-MAE of 0.1770 and the highest V-QWK of 0.8246.

\subsection{Ablation Studies}
\noindent\textbf{Backbone Scale.}
We first study the effect of backbone scale within the Qwen3.5 family. We evaluate 5 backbone variants from Qwen3.5-0.8B to Qwen3.5-27B using the same training protocol. The results are shown in~\cref{tab:backbone_scale}. Scaling from Qwen3.5-0.8B to Qwen3.5-4B consistently reduces both S-MAE and V-MAE. Further scaling yields only small changes across the 4 metrics. Qwen3.5-9B reduces S-MAE by only 0.0018 over Qwen3.5-4B, while its inference time increases from 2.80~s to 4.18~s. Qwen3.5-27B provides no further reduction in S-MAE. This suggests that additional model capacity offers diminishing returns for the structured rating task under the same training data and supervision. We therefore use Qwen3.5-4B as the default backbone.

\begin{table}[t]
    \centering
    \caption{Ablation on backbone scale.}
    \label{tab:backbone_scale}
    \small
    \resizebox{\linewidth}{!}{
    \begin{tabular}{rccccc}
        \toprule
        Model & S-MAE & S-QWK & V-MAE & V-QWK & Time(s) \\
        \midrule
        0.8B & 0.1589 & 0.8538 & 0.1964 & 0.8079 & 1.31 \\
        2B   & 0.1456 & 0.8657 & 0.1856 & 0.8208 & 1.80 \\
        4B   & 0.1388 & 0.8645 & 0.1770 & 0.8246 & 2.80 \\
        9B   & 0.1370 & 0.8567 & 0.1724 & 0.8366 & 4.18 \\
        27B  & 0.1408 & 0.8588 & 0.1760 & 0.8309 & 6.40 \\
        \bottomrule
    \end{tabular}
    }
    \vspace{-3mm}
\end{table}

\noindent\textbf{Training Data Scale.} 
We vary the training set from 25\% to 100\% while keeping the validation and test sets fixed. The subsets are constructed at the object level and are nested across data scales. The results are shown in~\cref{tab:data_scale}. Increasing the training data consistently reduces both S-MAE and V-MAE. S-MAE decreases from 0.2091 at 25\% data to 0.1388 at 100\%, a relative reduction of 33.6\%. V-MAE similarly decreases from 0.2523 to 0.1770, corresponding to a 29.8\% reduction. Most of the improvement occurs between 25\% and 50\%, with smaller gains at larger scales.

\begin{table}[t]
    \centering
    \caption{Ablation on training data scale.}
    \label{tab:data_scale}
    \small
    \begin{tabular}{rcccc}
        \toprule
        Data & S-MAE & S-QWK & V-MAE & V-QWK \\
        \midrule
        25\%  & 0.2091 & 0.7851 & 0.2523 & 0.7493 \\
        50\%  & 0.1467 & 0.8721 & 0.1847 & 0.8256 \\
        75\%  & 0.1462 & 0.8644 & 0.1830 & 0.8168 \\
        100\% & 0.1388 & 0.8645 & 0.1770 & 0.8246 \\
        \bottomrule
    \end{tabular}
    \vspace{-3mm}
\end{table}

\noindent\textbf{Prompting and Fine-tuning.}
Finally, we compare zero-shot, one-shot, three-shot, and LoRA SFT using the same Qwen3.5-4B backbone. Zero-shot uses no examples, while one-shot and three-shot use 1 and 3 fixed examples from the training set, respectively. Each example contains 5 rendered views and the corresponding structured response. The examples are selected to cover different label counts and quality levels. As shown in~\cref{tab:supervision_strategy}, one-shot increases S-MAE from 0.4013 to 0.6292 and decreases S-QWK from 0.5441 to 0.4906 compared with zero-shot, indicating that a single demonstration does not necessarily improve rating accuracy. Three-shot improves S-QWK to 0.5994, while S-MAE remains close to zero-shot at 0.4088, showing greater improvement in ordinal agreement than in continuous score accuracy. LoRA SFT achieves an S-MAE of 0.1388 and an S-QWK of 0.8645, corresponding to a 66.0\% reduction in S-MAE and an S-QWK gain of 0.2651 over three-shot. At the view level, LoRA SFT reduces V-MAE by 62.6\% and improves V-QWK by 0.2815. These results show that fine-tuning on human ratings outperforms few-shot prompting.

\begin{table}[t]
    \centering
    \caption{Comparison of prompting and fine-tuning.}
    \label{tab:supervision_strategy}
    \small
    \begin{tabular}{rcccc}
        \toprule
        Setting & S-MAE & S-QWK & V-MAE & V-QWK \\
        \midrule
        Zero-shot  & 0.4013 & 0.5441 & 0.4819 & 0.5194 \\
        One-shot   & 0.6292 & 0.4906 & 0.6510 & 0.4825 \\
        Three-shot & 0.4088 & 0.5994 & 0.4734 & 0.5431 \\
        \midrule
        LoRA SFT & \textbf{0.1388} & \textbf{0.8645} & \textbf{0.1770} & \textbf{0.8246} \\
        \bottomrule
    \end{tabular}
    \vspace{-3mm}
\end{table}

\subsection{User Study}
Our user study addressed 3 research questions about \textit{agreement between model and human ratings} (RQ1), \textit{agreement between model and human rankings} (RQ2), and \textit{user acceptance of model rankings} (RQ3). The study was approved by the IRB of our university\footnote{Detailed IRB information is omitted for anonymous review and will be disclosed upon publication.}. All participants provided written informed consent before the study.

\noindent\textbf{Participants and Setup.}
We recruited 24 participants through a university forum (12 male and 12 female, aged 19--34), all with normal or corrected-to-normal vision. Participants completed the study through a custom web-based questionnaire on their own computers and were instructed to adjust browser zoom for clear viewing.

\noindent\textbf{Samples and Tasks.}
We selected 30 annotation layouts from 10 objects in the ALQA test set, with 3 layouts per object and 5 rendered views per layout. The study included 3 tasks: (1) Quality Rating (RQ1). Participants rated 10 selected layouts across 5 viewpoints and 5 quality dimensions using a 1--5 scale. (2) Quality Ranking (RQ2). Participants ranked the 3 layouts of each object from best to worst without seeing the model rankings or allowing ties. (3) Order Checking (RQ3). Participants inspected the model rankings for the same 10 objects and adjusted them when necessary.

\noindent\textbf{Study Design.}
The experiment used a 2-group design with 12 participants in each group (6 male and 6 female). One group completed Quality Rating followed by Quality Ranking, while the other completed Order Checking. We selected 5 objects with small differences in sample-level human reference scores among their layouts and 5 with larger differences. After ordering each object's layouts by score, the 2 sets of objects had mean minimum gaps between adjacent layouts of 0.17 and 0.55 and mean score ranges of 0.50 and 1.28, respectively. The selected objects covered 6 categories and contained an average of 12.7 labels, ranging from 7 to 19. For Quality Rating, we selected 1 layout from each object, including 3 highest-ranked, 4 middle-ranked, and 3 lowest-ranked layouts based on human reference scores. Their reference scores ranged from 2.60 to 4.70. Quality Ranking and Order Checking used all 3 layouts of each object. Each task was completed by 12 participants with 10 trials each, yielding 360 trials in total. We also recorded task completion times. For Quality Rating, we computed Spearman's $\rho$ between mean participant ratings and model ratings across the 10 layouts. For Quality Ranking, we counted matching layout pairs between human and model rankings. For Order Checking, we counted reversed layout pairs after participants reviewed the model rankings. For Quality Ranking and Order Checking, we averaged the metrics computed for individual participants.

\begin{figure}[!t]
    \centering
    \includegraphics[width=0.95\columnwidth]{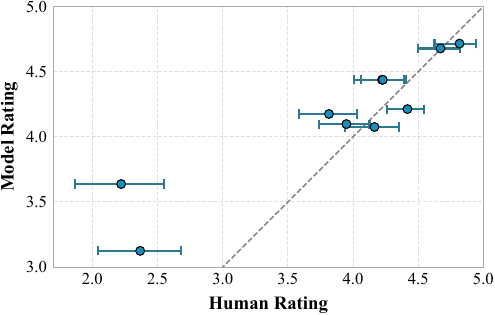}
    \caption{Mean participant ratings versus model ratings in the Quality Rating task. Each point represents one layout, and the dashed line indicates equal ratings. Horizontal error bars indicate 95\% bootstrap confidence intervals for mean participant ratings.}
    \label{fig:user_rating}
    \vspace{-3mm}
\end{figure}

\begin{figure}[!t]
    \centering
    \includegraphics[width=0.95\columnwidth]{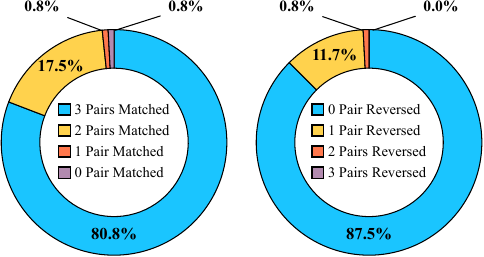}
    \caption{Trial distributions by the number of layout pairs matched in Quality Ranking (left) and reversed in Order Checking (right).}
    \label{fig:user_ranking}
    \vspace{-3mm}
\end{figure}

\noindent\textbf{Results.}
For Quality Rating, \cref{fig:user_rating} compares mean participant ratings with model ratings across the 10 layouts. The ratings exhibit a strong positive correlation (Spearman's $\rho=0.900$, $p<0.001$), although the model overestimates the quality of 2 low-rated layouts. For Quality Ranking, \cref{fig:user_ranking} (left) shows the agreement between human and model rankings. Participants' rankings fully agreed with the model rankings in 80.8\% of the trials, while 17.5\% had 2 matching layout pairs and only 1.7\% had fewer than 2 matching pairs. The overall pairwise agreement reached 92.8\%, indicating strong agreement between human and model rankings. For Order Checking, 
\cref{fig:user_ranking} (right) presents the revisions made by participants to the model rankings. Participants retained the original rankings in 87.5\% of the trials, while 11.7\% involved 1 reversed pair and 0.8\% involved 2 reversed pairs. No trials involved 3 reversed pairs. These results suggest that participants generally accepted the model rankings, with only 1 pair reversed in most revisions.

\section{Conclusion}
In this paper, we propose an MLLM-based method for automatic multi-view quality assessment of 3D annotation layouts. Our method learns from human ratings to predict a rating for each view and quality dimension, with the predictions aggregated into view-level and sample-level quality scores. We also construct ALQA to support the training and evaluation of automatic layout quality assessment methods. The experiments demonstrate that our method produces quality scores that closely reflect human judgments. These scores support comparisons between alternative layouts.


{
    \small
    \bibliographystyle{ieeenat_fullname}
    \bibliography{mycvpr}
}


\end{document}